\documentclass{article}

\usepackage{PRIMEarxiv}
\usepackage{longtable}
\usepackage{booktabs}
\usepackage{amsmath,amssymb,amsfonts}
\usepackage{amsthm}
\usepackage{mathrsfs}
\usepackage[title]{appendix}
\usepackage{xcolor}
\usepackage{textcomp}
\usepackage{manyfoot}
\usepackage{booktabs}
\usepackage{listings}
\usepackage{graphicx}
\usepackage{lineno}
\usepackage{url}
\usepackage{times}
\usepackage{wrapfig}
\usepackage{textgreek}
\usepackage{dsfont}
\usepackage{verbatim}
\usepackage{bm}
\usepackage[]{algorithm2e}
\usepackage{colortbl}
\usepackage{enumerate}
\usepackage{tikz}
\usepackage{tablefootnote}
\usepackage{booktabs, multirow} 
\usepackage{soul} 
\usepackage{hyperref}
\usepackage{stfloats}
\usepackage{enumitem}
\usepackage{array}
\usepackage{algorithmic}
\usepackage{caption}
\usepackage{subcaption}
\usepackage{graphicx}
\usepackage{wrapfig}
\usepackage{tabularx}

\makeatletter
\g@addto@macro{\endtabular}{\rowfont{}}
\makeatother
\newcommand{\rowfonttype}{}
\newcommand{\rowfont}[1]{
   \gdef\rowfonttype{#1}#1%
}
\SetKwInput{kwInit}{Init}

\definecolor{berrin}{RGB}{0, 0, 0}
\definecolor{mehmet}{RGB}{36, 113, 163}

\newcommand{\bx}{\bm{x}}
\newcommand{\bmu}{\bm{\mu}}
\newcommand{\bsigma}{\bm{\sigma}}
\newcommand{\bz}{\bm{z}}

\newcommand{\bzero}{\bm{0}}
\newcommand{\bI}{\bm{I}}

\usepackage{textcomp}
\usepackage{xcolor}
\makeatletter
\newcommand\notsotiny{\@setfontsize\notsotiny{7.5}{8}}
\newcommand\notsofootnote{\@setfontsize\notsotiny{8.75}{9.25}}
\newcommand\notsosmall{\@setfontsize\notsotiny{9.75}{10.25}}
\makeatother
\usepackage[utf8]{inputenc} 
\usepackage[T1]{fontenc}    
\usepackage{hyperref}       
\usepackage{url}            
\usepackage{booktabs}       
\usepackage{amsfonts}       
\usepackage{nicefrac}       
\usepackage{microtype}      
\usepackage{lipsum}
\usepackage{fancyhdr}       
\usepackage{graphicx}       

\title{Variational Self-Supervised Contrastive Learning Using Beta Divergence} 

\author{
  Mehmet Can Yavuz, Berrin Yanikoglu \\
  Sabanci University, \\
  34956, Istanbul \\
  Türkiye \\
  \texttt{\{berrin, mehmetyavuz\}@sabanciuniv.edu} \\
}

\begin{document}
\maketitle

\begin{abstract}
Learning a discriminative semantic space using unlabelled and noisy data remains unaddressed in a multi-label setting. We present a contrastive self-supervised learning method which is robust to data noise, grounded in the domain of variational methods. The method (VCL)  utilizes variational contrastive learning with beta-divergence to learn robustly from unlabelled datasets, including uncurated and noisy datasets. We demonstrate the effectiveness of the proposed method through rigorous experiments including linear evaluation and fine-tuning scenarios with multi-label datasets in the face understanding domain. In almost all tested scenarios, VCL surpasses the performance of state-of-the-art self-supervised methods, achieving a noteworthy increase in accuracy.
\end{abstract}


\section{Introduction}

Supervised deep learning approaches require large amounts of labelled data. While transfer learning with pretrained models is commonly used for addressing the labelled data shortage, the recent research focus has been on unsupervised and especially self-supervised  methods  trained with unlabelled or weakly labelled data that may be collected from the web \cite{tian2021divide, goyal2021self, zhong2022self, cole2022does}. Even though it is easy to have access to  uncurated data sets collected from the web, these data sets often lack useful labels, making it necessary to develop robust learning algorithms to enhance the performance of a visual classifier \cite{li2017webvision, yavuz2021yfcc}.

Self-supervised learning utilizes supervisory signals that are generated internally from the data, eliminating the need for external labels. A significant part of self-supervised learning research involves using \textit{pretext tasks} to learn embedding representations that are helpful in downstream tasks. Pretext tasks may involve patch context prediction \cite{mundhenk2018improvements}; solving jigsaw puzzles from the same \cite{noroozi2018boosting}; colorizing images \cite{larsson2017colorization,zhang2016colorful}; predicting noise \cite{bojanowski2017unsupervised}; counting \cite{noroozi2017representation}; inpainting patches \cite{pathak2016context};  spotting artifacts \cite{jenni2018self}; generating images \cite{ren2018cross}; predictive coding \cite{van2018representation}; or instance discrimination \cite{wu2018unsupervised}. 
\textcolor{black}{Self-supervised learning methods aim to learn embedding vectors that effectively represent the underlying semantic relationships among the data.}

State-of-the-art self-supervised literature also includes approaches that have been proposed for highly curated sets such as the Imagenet \cite{deng2009imagenet}, namely  SimCLR \cite{chen2020simple}, BYOL \cite{grill2020bootstrap}, NNCLR \cite{dwibedi2021little}, VICReg \cite{bardes2021vicreg}, Barlow Twins \cite{zbontar2021barlow}, MoCo \cite{chen2021empirical} and TiCo \cite{zhu2022tico}. A significant number of these approaches uses the \textit{contrastive learning} paradigm that aims to minimize the distance between embeddings of similar (positive) samples generated from various random transformations or augmentations of the input image, while simultaneously increasing the distance between non-similar (negative) samples.
The variational contrastive representation learning has only recently been studied in semi-supervised fashion \cite{vcl}.

In the realm of noisy data, there is a limited amount of research. Some approaches utilize standard clustering algorithms capable of handling outlier noise in semantic space \cite{zheltonozhskii2022contrast, tian2021divide}, while others enforce neighbor consistency to address the issue \cite{iscen2022learning}. 

In addition to addressing noisy data, there is an upcoming body of research on \textit{continual learning} that inherently tackles the issue of noise \cite{karim2022cnll}. Furthermore, the literature also explores the use of multi-stage algorithms for managing noisy data effectively \cite{smart2023bootstrapping}. \textcolor{black}{The main limitation of these methodologies is that there always may not an appropriate label for noisy samples originating from web-collected datasets. Furthermore, these approaches largely rely on initial self-supervised pre-training, which is the focus of our study.}

The aim of this work is to enhance the pre-training efficiency of classifiers via employing robust self-supervised techniques, by harnessing an almost infinite source of internet collected images. While abundant, such images are unlabelled and noisy. We propose a self-supervised variational contrastive method in combination with the beta divergence formulation that is robust to data noise. The variational approach is first suggested for auto-encoders and recently used in conjunction with contrastive learning in a semi-supervised paradigm in \cite{kingma2013auto,vcl}.

The proposed approach of VCL is validated across a range of diverse settings, including face attribute learning with the medium-sized CelebA and YFCC-CelebA datasets. Overall, the proposed approach exhibited promising results across a variety of settings and scenarios.

\section{Method}
Building upon our previous work that extended the contrastive learning framework with a variational approach and demonstrated significant improvements in accuracy within a \textit{semi-supervised} learning context, we now present a \textit{self-supervised} variational contrastive learning algorithm, combined with the beta divergence to make it suitable for handling noisy data \cite{vcl}.

\begin{figure}[thb]
  \begin{center}
    \includegraphics[width=0.9\textwidth]{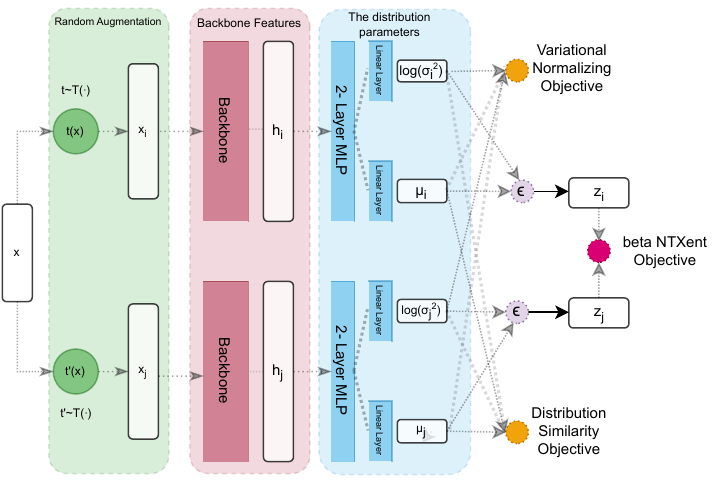}
  \end{center}
  \caption{Diagram of  our proposed model from left to right. First two augmentations $x_i$ and $x_j$ are obtained from the input and their latent vectors ($h_i$ and $h_j$) are extracted. Then the Gaussian sampling head learns the distribution parameters (mean and log variance) and samples from the learned distribution. Light color boxes indicate the same operation or the shared weights.}
  \label{Fig:VariationalContrastive}
\end{figure}

The approach first of all is a contrastive one, based on positive and negative samples. The generation of two augmentations of the input image is and the contrastive set up is explained in Sections 2.1 and 2.2 respectively. As in the original variational approach, a Gaussian sampling head learns the distribution of images in the embedding space, using an objective function with three terms, as explained in Sections 2.3 and 2.4. The whole system is depicted in Figure \ref{Fig:VariationalContrastive}. 

\subsection{The Contrastive Learning Framework}

In the contrastive learning process, we use a mini-batch of $N$ randomly selected images from the dataset. From each image in the mini-batch, two random augmentations are obtained (see Subsection \ref{subsection:augmentation}), resulting in a mini-batch of size $2N$ data points in total. For a positive pair (two augmentations of an image), the rest of the $2(N-1)$ samples within the mini-batch constitute the negative examples. Thus, within a mini-batch of size $N$, there are $N$ positive and $2(N-1)$ negative pairs.

Following the creation of the augmented images, the backbone network is employed to extract fixed-length embedding vectors for each of the two augmentations (shown as red box in Fig. 1). As in traditional contrastive learning, the objective function is designed to obtain similar embeddings for positive pairs, compared to the negative samples. 
This is achieved by minimizing the contrastive loss, which encourages the model to learn representations that are similar for positive pairs and dissimilar for negative pairs.

\subsection{Augmentations} 
\label{subsection:augmentation}

\begin{figure}[thb]
    \centering
    \includegraphics[width=0.75\linewidth]{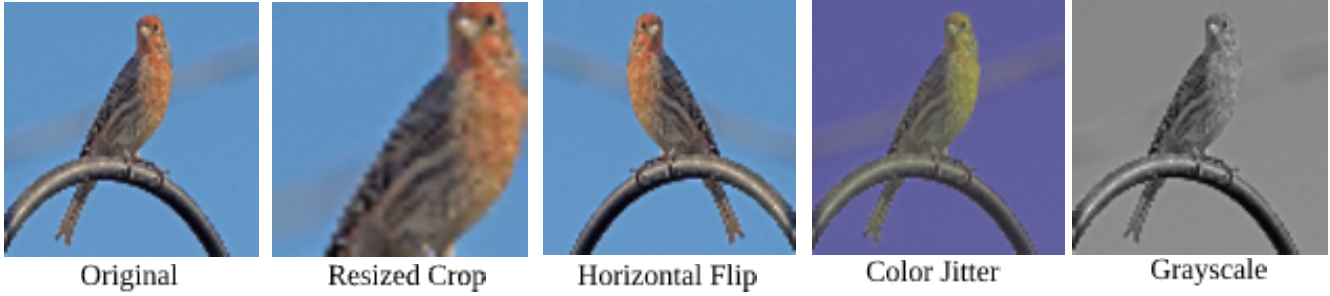}
    \caption{Four different augmentations for contrastive design.}
    \label{fig:augs}
\end{figure}

Data augmentation is a widely used technique to prevent overfitting in supervised deep learning systems. 
As illustrated in Figure \ref{Fig:VariationalContrastive}, a given an input $x$, the stochastic function $t(\cdot)$ is used to obtain two random augmentations of the input image, $\widetilde{x_i}$ and $\widetilde{x_j}$. These two correlated images constitute a positive pair, while the other examples in the mini-batch constitute the negative pairs.  We follow the augmentation strategies suggested in the original article \cite{chen2020simple}. Specifically, we use Resize Crop, Horizontal Flip, Grayscale, Color Jitter augmentations (see Figure \ref{fig:augs}).

\subsection{Feature Extraction and Gaussian Sampling}
\label{subsection:feature}

The backbone model $f(\cdot)$  extracts fixed-length embedding vectors for the two augmentations of the input image, indicated as $h_i = f(x_i)$, where $h_i \in R^d$.
\textcolor{black}{The Gaussian sampling head ($g_z$) is a projection head that compute a mean and a log-variance for the distributions of the two representations, allowing for the sampling of new representations from the learned distribution.} Specifically:
\begin{equation}
    \label{mapping}
    (\bmu, \log \bsigma^2) = g_{z}(f_{\theta}(t(\bx)))\\
\end{equation}
A sample $\bz$ is then drawn from this distribution, after the so called reparameterization trick, as in  \cite{kingma2013auto}:
\begin{equation}
    \label{parametrization}
    \bz = \bmu + \bsigma^2 \odot \epsilon
\end{equation}
where $\epsilon \sim \mathcal{N}(\bzero,\,\bI)$ and $\odot$ is the element-wise multiplication. The reparameterization  allows backpropagation to be used, in spite of the sampling process. 
The samples obtained from the two augmentations contribute to the loss term given in Eqs. \ref{eq:ntxent}, \ref{eq:equalizer}.

\subsection{Objective Function} 
\label{subsection:objective}

The training objective is to minimize the overall loss function that is composed of three terms:
i)  beta-NT-Xent loss term derived from the beta divergence; ii) Distribution Similarity Loss; and iii) Distribution Normalizing Loss,  explained below. 

\textbf{beta-NT-Xent loss} is used as the loss function that considers the similarity between an image embedding sample $z_i \sim p_{\theta}(Z_i|X_i)$ and another sample $z_j \sim p_{\theta}(Z_j|X_j)$, compared to the similarity between $z_i$ and the negatives samples $z_{j\neq k}$. The loss  for a positive embedding pairs $z_i, z_j$ is defined as the ratio between the similarity of positive and negative samples in the mini-batch:
\begin{equation}
    \label{eq:ntxent}
    l^{\beta}_{i,j} = -\log \frac{\exp(\beta dist(z_i, z_j) / \tau)}{\sum_{k=1}^{2N}\mathds{1}_{k \neq i}\exp(\beta dist(z_i, z_k) / \tau)}
\end{equation}
where $\tau$ is the temperature parameter, N is the batch size, and $\mathds{1}_{k\neq i}$ is the indicator function evaluating to 1 if and only if $k\neq i$ and $\beta dist$ is a similarity metric between two sample embeddings $z_i$ and $z_j$, such that:
\begin{equation}
    \beta dist(z_j, z_i) =\nonumber -\frac{\beta+1}{\beta} \left(\frac{1}{(2\pi\sigma^2)^{\beta/2}} exp\big(-\frac{\beta}{2\sigma^2} d(z_j, z_i) \big) - 1\right)
    \label{eq:beta}
\end{equation}
where $\sigma$ is the standard deviation of the hypothesized Gaussian distribution which is set to 0.5 as suggested by \cite{akrami2022robust}, $\beta$ is a hyper-parameter which is set to X, and $d$ function is based on Euclidean distance between two vectors $z_j, z_i$ in semantic space. 

\textbf{Distribution Similarity Loss.} The second loss term encourages that two augmentations ($x_i$, $x_j$) of the input should be  drawn from similar distributions, $q_1$ and $q_2$. The loss term penalizes the Jensen–Shannon divergence between the two distributions \cite{Odaibo2019TutorialDT}. Using Gaussian distributions, the Jensen-Shannon divergence reduces to:
\begin{eqnarray}
    \label{eq:equalizer}
    &&l^{dist}_{i,j} = - \frac{1}{2} \bigg[(\log(\sigma_{i}) - \log(\sigma_{j}))\bigg] + \frac{1}{4} \bigg[ \frac{(\mu_{i}-\mu_{m})^2 + (\mu_{j}-\mu_{m})^2}{\sigma^2_{m}}\bigg]
\end{eqnarray}

where 
$\log(\sigma_{i/j})$ and $\mu_{i/j}$ are the outputs of the variational encoder. The $\mu_{m}$ and $\sigma_{m}$ are the means of the two means and two standard deviations, respectively.

\textbf{Distribution Normalizing Loss.} The third loss term encourages the learned distributions to be standard Gaussian \cite{Odaibo2019TutorialDT}, to eliminate degenerate solutions:
\begin{align}
    \label{eq:normalizer}
    l^{norm}_{i} &=  D_{KL}(q_\theta(z_{i}|x_{i}) || \mathcal{N}(0,1)) = - \frac{1}{2}\left[1 +\log(\sigma^2_{i}) - \sigma^2_{i} - \mu^2_{i} \right] 
\end{align}
where $\log(\sigma)$ and $\mu$ are the outputs of the variational encoder and directly computed to minimize.

\textbf{Overall Objective Function.} 
The optimization is processed based on the total loss, which is comprised of the four loss terms defined in Equations 3-5:
\begin{eqnarray}
    \mathcal{L}_{total} = \frac{1}{N}\sum^{N}_{i=1} \bigg\{l^{norm}_{i} + \sum_{j=1}^{2N} \mathbf{1}_{i \neq j} (l^{\beta}_{i,j} + l^{dist}_{i,j})\bigg\}
    \label{eq:totalloss}
\end{eqnarray}
where $N$ is the number of images in the mini-batch and $\{2k-1,2k\}$ notation follows \cite{chen2020simple}. The reader can refer to Algorithm \ref{alg:vcl} for a detailed explanation.

\begin{algorithm}[htb]
    \caption{Pseudo-code for Variational Contrastive Learning with Beta Divergence.}
    \label{alg:vcl}
    \SetAlgoLined
    \KwResult{Optimized parameters of the base encoder $f(\mathbin{\cdot})$}
    \textbf{Input:} Batch size $N$, temperature parameter $\tau$, beta parameter $\beta$, augmentation function $T(\mathbin{\cdot})$, encoder $f(\mathbin{\cdot})$, Gaussian sampling head $g(\mathbin{\cdot})$\\
    \BlankLine
    \For{each minibatch of $N$ examples $\{x_i\}_{i=1}^{N}$}{
        \tcp{Generate augmented views of each example}
        Compute $x_{i}^{'} = T(x_i)$ and $x_{i}^{''} = T(x_i)$\\
        \BlankLine
        \tcp{Compute representations of the augmented views}
        Compute $h_{i}^{'} = f(x_{i}^{'})$ and $h_{i}^{''} = f(x_{i}^{''})$\\
        \BlankLine
        \tcp{Learn the distribution parameters and sample from the learned distribution}
        Sample $z_{i}^{'} = g(h_{i}^{'})$ from $N(\mu_i^{'}, \sigma_i^{'})$\\
        Sample $z_{i}^{''} = g(h_{i}^{''})$ from $N(\mu_i^{''}, \sigma_i^{''})$\\
        \BlankLine
        \tcp{Compute the beta NT-Xent and variational objectives}
        Compute $\mathcal{L}_i = l^{norm}_{i} + \sum_{j=1}^{2N} \mathbf{1}_{i \neq j} (l^{\beta}_{i,j} + l^{dist}_{i,j})$
        \BlankLine
        \tcp{Back-propagation and parameter update}
        Update the parameters of $f(\mathbin{\cdot})$ and $g(\mathbin{\cdot})$ by minimizing $\mathcal{L}$\\
    }
\end{algorithm}

We considered three  extensions of the Kullback-Leibler divergence that  differ in sensitivity and robustness, namely alpha, beta, and gamma divergences. 
We chose beta divergence which is often preferred for practical algorithms such as robust PCA and clustering, robust ICA,  robust NMF and robust VAE due to its balanced traits \cite{mollah2010robust, mollah2006exploring, kompass2007generalized, akrami2022robust}.

\subsection{Implementation Details}
We used the VGG16bn backbone which is often used as a benchmark in the face recognition community with an embedding dimension of 4096 \cite{ahmed2021relative}.  For training, we used the AdamW optimizer with learning rate of 1e-3 and weight decay of 0.01, together with the Cosine Annihilation Scheduler. The batch size for all methods was 128. We trained the networks with CelebA for 800 epochs (1M iterations) and YFCC-CelebA for 500 epochs (1.5M iterations).

For augmentations, we used the following ranges: resizing (scale between [0.2, 1.0]), cropping (128 random crops), grayscale transformation (with probability 0.2), and color jitter (with probability 0.8,  brightness in [0.6,1.4], contrast in [0.6,1.4], saturation in [0.6,1.4] and hue  in [0.9,1.1]).  We optimized the hyper-parameter beta for different values using grid search and found out the optimal temperature as 0.07 and beta as 0.005 for CelebA and YFCC-CelebA datasets.
\footnote{The code is available at \href{https://github.com/verimsu/VCL}{https://github.com/verimsu/VCL}}

\section{Experimental Evaluation}
\label{section:experiments}

We evaluate the proposed self-supervised algorithm first with two protocols (linear and low-shot) that are widely accepted in the self-supervised  learning literature. In both protocols, the self-supervised backbone network is extended by adding a single, fully connected layer that is trained for the corresponding classification task using the learned representations. In a third protocol, we evaluate the usefulness of the learned features.
\begin{enumerate}[label=(\roman*)]
    \item In the \textit{linear evaluation protocol} used in the literature, we  add a single layer as classification head and train only that layer with the labeled dataset, while the rest of the network is kept unchanged. The aim of this evaluation is to compare the quality of the learned representations to those obtained by fully supervised training.
    \item In the \textit{low-shot evaluation protocol}, we perform the self-supervised training without using labels and only use a small subset of the training data for fine-tuning all layers of this network. Here, the aim is to demonstrate the representation learning effectiveness of the proposed algorithm for low-data regimes, simulated by using only a small percentage of the data labels. In other words, the inquiry pertains to whether the enhancement of classifier efficiency is attainable through the utilization of web-sourced datasets.
\end{enumerate}

In all settings, we  compare our algorithms with state-of-the-art including SimCLR \cite{chen2020simple}, BYOL \cite{grill2020bootstrap}, NNCLR \cite{dwibedi2021little}, MoCo \cite{chen2021empirical}, VICReg \cite{bardes2021vicreg}, TiCo \cite{zhu2022tico}, and Barlow Twins \cite{zbontar2021barlow}.

Information about the datasets used in the evaluations is summarized in Table \ref{tab:statistics}. CelebA is a well-known public datasets used for face attribute recognition and ranking, where each photograph is labelled in terms of 40 attributes.  Finally,  YFCC-CelebA is a subset of the web-collected YFCC, where images matching  attributes in CelebA or their opposites (if they exist) were selected \cite{yavuz2021yfcc}. This dataset is thus weakly labelled and noisy, with samples given in Fig. \ref{fig:case1_meanvecs}.

\begin{table}[thb]
\centering
\caption{Summary of the datasets used in the study.}

\begin{tabular}{l|c|c|c|c|c}\midrule
Datasets&Setting &Classes &Train &Validation &Test \\\midrule
\textit{Medium-Sized} & & & & &\\
CelebA \cite{liu2015faceattributes}& Multi-labelled &40 &162,770 &19,867 &19,962 \\
YFCC-CelebA \cite{yavuz2021yfcc}& Multi/Weakly labelled &59 &392,220 &- &- \\
\bottomrule
\end{tabular}
\label{tab:statistics}
\end{table}

\begin{table}[thb]\centering
\caption{Comparison of model accuracies using the entire labeled CelebA dataset for linear evaluation. Bold indicates the best results; underline indicates the second best. }

\begin{tabular}{l|cc|c}\toprule
Multi-label                 &Supervised     & Self-supervised  & Mean Acc. on\\
Logistic Reg. Evaluation    &Pre-training   & Training          & CelebA\\
                             &w. Imagenet   & w. CelebA         & Test Set \\\midrule
\textit{Transfer Learning w. VGG16bn} &yes &no & 86.31\% \\\midrule
\textit{Self-supervised methods} & & & \\
Barlow Twins \cite{zbontar2021barlow} &no &yes & 87.69\% \\
BYOL \cite{grill2020bootstrap} &no &yes & 86.78\% \\
MoCo \cite{chen2021empirical} &no &yes & 87.92\% \\
NNCLR \cite{dwibedi2021little} &no &yes & 86.51\% \\
SimCLR \cite{chen2020simple} &no &yes & \underline{87.98\%} \\
TiCo \cite{zhu2022tico} &no &yes &\underline{87.98\%} \\
VICReg \cite{bardes2021vicreg} &no &yes &87.44\% \\\midrule
{This work} - VCL &no &yes &89.20\% \\
This work - VCL (beta div.)&no &yes & \textbf{89.23}\% \\
\bottomrule
\end{tabular}
\label{table:celeba}
\end{table}
\subsection{Evaluation with Multi-Label CelebA Dataset}

The effectiveness of the proposed method for a multi-label problem has been assessed using the CelebA dataset, which is widely recognized in the field of face attribute recognition. While facial attributes have been previously studied using supervised deep learning systems \cite{liu2015faceattributes, sharif2014cnn, song2014exploiting, zhu2014multi, rozsa2016facial, zhong2016face}, face attribute recognition has not been well studied in the context of self-supervised learning \cite{aly2018multi, 8756609,2210.03853,1808.06882}.

In this experiment, we applied self-supervised training on the training portion of the CelebA dataset \textit{without} using the labels. Then the evaluation of the learned representations is carried out in accordance with  protocol (i). 
In other words, we have added a linear layer that takes the learned representations as input and is trained with the whole labeled dataset, while the rest of the network is kept unchanged.
The results are presented in Table \ref{table:celeba}. 
The first row indicates the results obtained with transfer learning with the VGG16bn network trained with the ImageNet dataset and used as feature extractor prior to the linear layer. While this approach is preferred especially with smaller labelled datasets, it also obtains the lowest accuracy with 86.31\%.
Next, we evaluated some well-known self-supervised methods from the literature, which 
achived 86.51\% to 87.98\% accuracies, with best results  obtained by SimCLR \cite{kinakh2021scatsimclr} and TiCo methods \cite{zhu2022tico}.

The proposed method outperforms transfer learning by almost 3\% points (86.31 vs 89.23\%) and other self-supervised methods by over 1\% point (87.98 vs 89.23).
For this dataset, VCL beta-divergence (proposed) and VCL (evaluated as ablation) achieve almost the same performance. On the other hand, using the beta divergence results in clearly better performance when self-supervised learning is done with the noisy YFCC-CelebA dataset, as described in Section \ref{sec:yfcc}

\subsection{Evaluation with Web-Collected YFCC-CelebA Dataset}
\label{sec:yfcc}

\begin{figure}[thb]
\centering
\includegraphics[width=0.9\linewidth]{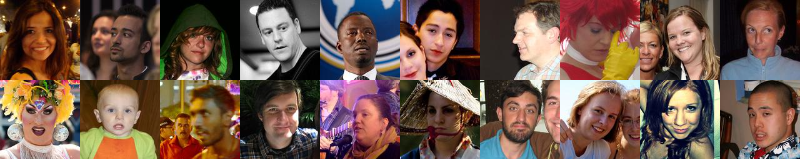}
\caption{Random samples from the YFCC-CelebA dataset \cite{yavuz2021yfcc}}.
\label{fig:case1_meanvecs}
\end{figure}

In a bid to further examine the robustness of our model in the face of data noise, we carried out a second experiment. For this, we implemented self-supervised training using the web-collected YFCC-CelebA dataset described in \cite{yavuz2021yfcc} and previously used in \cite{vcl} within a  semi-supervised learning paradigm. This dataset contains many non-face images and wrong labels, as shown in Figure \ref{fig:case1_meanvecs}.  This experiment was evaluated following protocol (ii). In other words, networks that are pretrained using the YFCC-CelebA dataset are fine-tuned within a low-data regime using 10\% or 1\% of the labeled CelebA dataset.

Table \ref{table:yfcc} compares the results of various well-known alternatives from the literature and the proposed model. Transfer learning with a  model trained only with supervised learning  with the ImageNet dataset (no  self-supervised pre-training) achieved the lowest accuracies, as was the case in the previous experiment. Self-supervised and semi-supervised methods from the literature achieved similar results for the 10\% labelled data case, while semi-supervised methods achieved better results for the very low-data regime. The proposed VCL model outperformed all other methods, achieving 91.01\% and 88.12\% accuracies when trained with 10\% or 1\% labeled data respectively, which corresponds to more than 1\% point increase over alternatives. 

Even though the two evaluation goals are different, we see that pre-training with the web-collected data and fine-tuning with 10\% of labelled CelebA results in higher accuracy (91.01\% in Table \ref{table:yfcc}), compared to pre-training with CelebA and using 100\% of labelled CelebA (89.23\% in Table \ref{table:celeba}).
Finally, when comparing VCL and VCL with beta divergence, we see that the latter is indeed more effective for this noisy dataset. 

\begin{table}[thb]\centering
\caption{Comparison of model accuracies using the unlabeled YFCC-CelebA dataset for pre-training. The models are then fine-tuned using 10\% and 1\% of the labeled CelebA dataset. Bold indicates the best results; underline indicates the second best.}

\begin{tabular}{l|cc|c|c}\toprule
                    & Supervised & Self-sup.     & Mean Acc. on              & Mean Acc. on  \\
Multi-label Model& Pre-trained & Training w.     & CelebA Test Set           & CelebA Test Set  \\
Fine-tuning Evaluation&w. Imagenet & YFCC-CelebA & - CelebA (10\%) &  - CelebA (1\%)\\\midrule 
\textit{Transfer Learning}&yes &no & 89.05\% & 86.34\% \\\midrule
\textit{Self-supervised methods} & & & & \\
Barlow Twins \cite{zbontar2021barlow} &no &yes & 89.18\% & 86.64\%\\
BYOL \cite{grill2020bootstrap} &no &yes & 89.40\% & 86.98\%\\
MoCo \cite{chen2021empirical} &no &yes & 89.83\% & 86.66\%\\
NNCLR \cite{dwibedi2021little} &no &yes & 89.95\% & 86.94\%\\
SimCLR \cite{chen2020simple} &no &yes & 89.89\% & 86.87\%\\
TiCo \cite{zhu2022tico} &no &yes &89.24\% & 86.85\%\\
VICReg \cite{bardes2021vicreg} &no &yes &\underline{89.97\%} & \underline{86.99\%}\\
\midrule
\textit{Semi-supervised} & & & & \\
CL-PL \cite{vcl} &yes & yes &89.43\% & 87.69\%\\
VCL-PL \cite{vcl} &yes &yes &89.68\% & 88.12\%\\
\midrule
{This work} - VCL &no &yes &90.15\% & 87.49\%\\
This work - VCL (beta div.)&no &yes & \textbf{91.01}\% & \textbf{88.12}\% \\

\bottomrule
\end{tabular}
\label{table:yfcc}
\end{table}

\section{Summary and Conclusions} 
\label{section:conc}
We proposed an algorithm in the family of contrastive learning framework, using beta-NT-Xent loss term derived from the beta divergence for robustness against outliers in the noisy set. The approach differs from simple contrastive design in its variational approach and use of beta divergence in the self-supervised objective. 

Our findings demonstrate that variational methods employing beta-divergence offer a robust alternative for tackling noisy datasets, multi-label settings and low-data regime. The results outperform existing self-supervised models especially for the case where the self-supervised learning is accomplished with a noisy dataset.

We also compared the effect of different objective components on accuracy, as well as temperature and beta value (not included for clarity). The combination of all components yielded the highest accuracies.

\section*{Acknowledgments}
This study is supported by a TÜBİTAK grant (Project no: 119E429). The numerical calculations reported in this paper are  performed at TUBITAK ULAKBIM, High Performance and Grid Computing Center (TRUBA resources). 

\begin{figure}[htbp]
    \centering
    
    \begin{subfigure}{0.475\textwidth}
        \includegraphics[width=\textwidth]{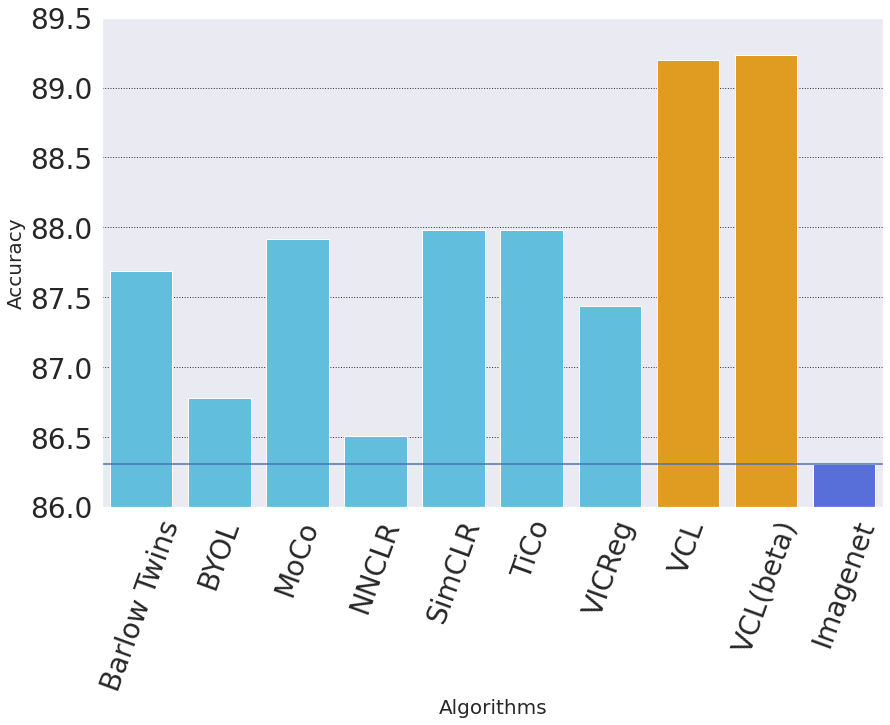}
        \caption{CelebA Linear Evaluation}
    \end{subfigure}
    \hfill
    \begin{subfigure}{0.475\textwidth}
        \includegraphics[width=\textwidth]{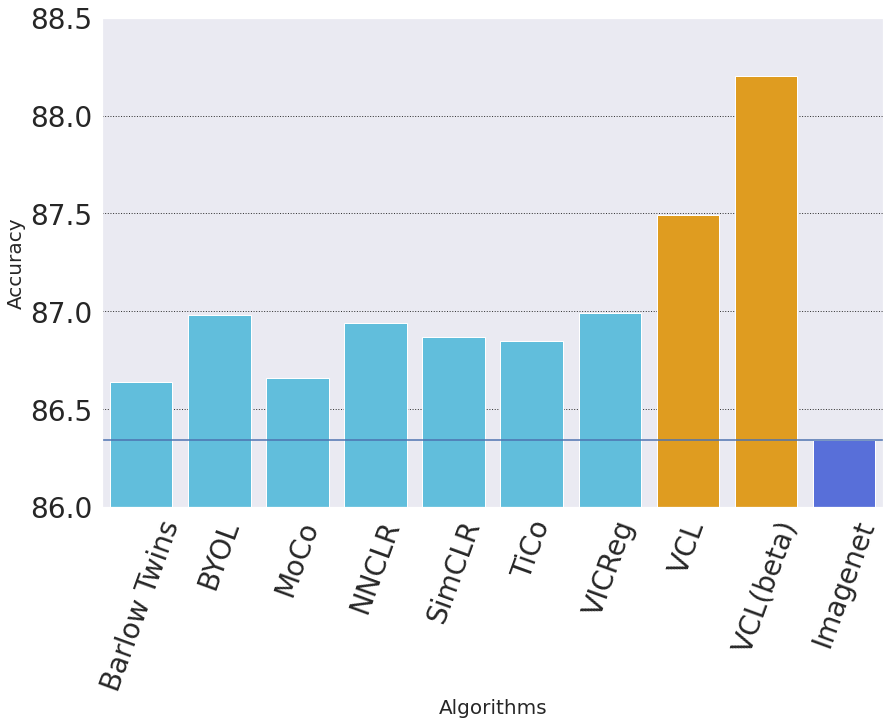}
        \caption{YFCC-CelebA Low Shot Ranking}
    \end{subfigure}
    
    \caption{Comparative performance of self-supervised pre-training models in multilabel tasks using CelebA and YFCC-CelebA datasets. The bar charts differentiate algorithms via color: orange represents VCL algorithms, turquoise represents state-of-the-art models, and blue bars denote supervised baselines.}
    \label{fig:att}
\end{figure}

\bibliographystyle{unsrt}  
\bibliography{main}  

\begin{thebibliography}{10}

\bibitem{tian2021divide}
Yonglong Tian, Olivier~J Henaff, and A{\"a}ron van~den Oord.
\newblock Divide and contrast: Self-supervised learning from uncurated data.
\newblock In {\em Proceedings of the IEEE/CVF International Conference on
  Computer Vision}, pages 10063--10074, 2021.

\bibitem{goyal2021self}
Priya Goyal, Mathilde Caron, Benjamin Lefaudeux, Min Xu, Pengchao Wang, Vivek
  Pai, Mannat Singh, Vitaliy Liptchinsky, Ishan Misra, Armand Joulin, et~al.
\newblock Self-supervised pretraining of visual features in the wild.
\newblock {\em arXiv preprint arXiv:2103.01988}, 2021.

\bibitem{zhong2022self}
Yuanyi Zhong, Haoran Tang, Junkun Chen, Jian Peng, and Yu-Xiong Wang.
\newblock Is self-supervised learning more robust than supervised learning?
\newblock {\em arXiv preprint arXiv:2206.05259}, 2022.

\bibitem{cole2022does}
Elijah Cole, Xuan Yang, Kimberly Wilber, Oisin Mac~Aodha, and Serge Belongie.
\newblock When does contrastive visual representation learning work?
\newblock In {\em Proceedings of the IEEE/CVF Conference on Computer Vision and
  Pattern Recognition}, pages 14755--14764, 2022.

\bibitem{li2017webvision}
Wen Li, Limin Wang, Wei Li, Eirikur Agustsson, and Luc Van~Gool.
\newblock Webvision database: Visual learning and understanding from web data.
\newblock {\em arXiv preprint arXiv:1708.02862}, 2017.

\bibitem{yavuz2021yfcc}
Mehmet~Can Yavuz, Sara~Atito Ali~Ahmed, Mehmet~Efe Kısaağa, Hasan Ocak, and
  Berrin Yanıkğlu.
\newblock Yfcc-celeba face attributes datasets.
\newblock In {\em 2021 29th Signal Processing and Communications Applications
  Conference (SIU)}, pages 1--4, 2021.

\bibitem{mundhenk2018improvements}
T~Nathan Mundhenk, Daniel Ho, and Barry~Y Chen.
\newblock Improvements to context based self-supervised learning.
\newblock In {\em Proceedings of the IEEE Conference on Computer Vision and
  Pattern Recognition}, pages 9339--9348, 2018.

\bibitem{noroozi2018boosting}
Mehdi Noroozi, Ananth Vinjimoor, Paolo Favaro, and Hamed Pirsiavash.
\newblock Boosting self-supervised learning via knowledge transfer.
\newblock In {\em Proceedings of the IEEE Conference on Computer Vision and
  Pattern Recognition}, pages 9359--9367, 2018.

\bibitem{larsson2017colorization}
Gustav Larsson, Michael Maire, and Gregory Shakhnarovich.
\newblock Colorization as a proxy task for visual understanding.
\newblock In {\em Proceedings of the IEEE conference on computer vision and
  pattern recognition}, pages 6874--6883, 2017.

\bibitem{zhang2016colorful}
Richard Zhang, Phillip Isola, and Alexei~A Efros.
\newblock Colorful image colorization.
\newblock In {\em European conference on computer vision}, pages 649--666.
  Springer, 2016.

\bibitem{bojanowski2017unsupervised}
Piotr Bojanowski and Armand Joulin.
\newblock Unsupervised learning by predicting noise.
\newblock In {\em International Conference on Machine Learning}, pages
  517--526. PMLR, 2017.

\bibitem{noroozi2017representation}
Mehdi Noroozi, Hamed Pirsiavash, and Paolo Favaro.
\newblock Representation learning by learning to count.
\newblock In {\em Proceedings of the IEEE International Conference on Computer
  Vision}, pages 5898--5906, 2017.

\bibitem{pathak2016context}
Deepak Pathak, Philipp Krahenbuhl, Jeff Donahue, Trevor Darrell, and Alexei~A
  Efros.
\newblock Context encoders: Feature learning by inpainting.
\newblock In {\em Proceedings of the IEEE conference on computer vision and
  pattern recognition}, pages 2536--2544, 2016.

\bibitem{jenni2018self}
Simon Jenni and Paolo Favaro.
\newblock Self-supervised feature learning by learning to spot artifacts.
\newblock In {\em Proceedings of the IEEE Conference on Computer Vision and
  Pattern Recognition}, pages 2733--2742, 2018.

\bibitem{ren2018cross}
Zhongzheng Ren and Yong~Jae Lee.
\newblock Cross-domain self-supervised multi-task feature learning using
  synthetic imagery.
\newblock In {\em Proceedings of the IEEE Conference on Computer Vision and
  Pattern Recognition}, pages 762--771, 2018.

\bibitem{van2018representation}
Aaron Van~den Oord, Yazhe Li, and Oriol Vinyals.
\newblock Representation learning with contrastive predictive coding.
\newblock {\em arXiv e-prints}, pages arXiv--1807, 2018.

\bibitem{wu2018unsupervised}
Zhirong Wu, Yuanjun Xiong, Stella~X Yu, and Dahua Lin.
\newblock Unsupervised feature learning via non-parametric instance
  discrimination.
\newblock In {\em Proceedings of the IEEE conference on computer vision and
  pattern recognition}, pages 3733--3742, 2018.

\bibitem{deng2009imagenet}
Jia Deng, Wei Dong, Richard Socher, Li-Jia Li, Kai Li, and Li~Fei-Fei.
\newblock Imagenet: A large-scale hierarchical image database.
\newblock In {\em 2009 IEEE conference on computer vision and pattern
  recognition}, pages 248--255. Ieee, 2009.

\bibitem{chen2020simple}
Ting Chen, Simon Kornblith, Mohammad Norouzi, and Geoffrey Hinton.
\newblock A simple framework for contrastive learning of visual
  representations.
\newblock In {\em International conference on machine learning}, pages
  1597--1607. PMLR, 2020.

\bibitem{grill2020bootstrap}
Jean-Bastien Grill, Florian Strub, Florent Altch{\'e}, Corentin Tallec, Pierre
  Richemond, Elena Buchatskaya, Carl Doersch, Bernardo Avila~Pires, Zhaohan
  Guo, Mohammad Gheshlaghi~Azar, et~al.
\newblock Bootstrap your own latent-a new approach to self-supervised learning.
\newblock {\em Advances in Neural Information Processing Systems},
  33:21271--21284, 2020.

\bibitem{dwibedi2021little}
Debidatta Dwibedi, Yusuf Aytar, Jonathan Tompson, Pierre Sermanet, and Andrew
  Zisserman.
\newblock With a little help from my friends: Nearest-neighbor contrastive
  learning of visual representations.
\newblock In {\em Proceedings of the IEEE/CVF International Conference on
  Computer Vision}, pages 9588--9597, 2021.

\bibitem{bardes2021vicreg}
Adrien Bardes, Jean Ponce, and Yann LeCun.
\newblock Vicreg: Variance-invariance-covariance regularization for
  self-supervised learning.
\newblock {\em arXiv preprint arXiv:2105.04906}, 2021.

\bibitem{zbontar2021barlow}
Jure Zbontar, Li~Jing, Ishan Misra, Yann LeCun, and St{\'e}phane Deny.
\newblock Barlow twins: Self-supervised learning via redundancy reduction.
\newblock In {\em International Conference on Machine Learning}, pages
  12310--12320. PMLR, 2021.

\bibitem{chen2021empirical}
Xinlei Chen, Saining Xie, and Kaiming He.
\newblock An empirical study of training self-supervised vision transformers.
\newblock In {\em Proceedings of the IEEE/CVF International Conference on
  Computer Vision}, pages 9640--9649, 2021.

\bibitem{zhu2022tico}
Jiachen Zhu, Rafael~M Moraes, Serkan Karakulak, Vlad Sobol, Alfredo Canziani,
  and Yann LeCun.
\newblock Tico: Transformation invariance and covariance contrast for
  self-supervised visual representation learning.
\newblock {\em arXiv preprint arXiv:2206.10698}, 2022.

\bibitem{vcl}
Mehmet~Can Yavuz and Berrin Yanikoglu.
\newblock \mbox{VCL-PL}: semi-supervised learning from noisy web data with
  variational contrastive learning.
\newblock In {\em 2022 26th International Conference on Pattern Recognition
  (ICPR)}, pages 740--747. IEEE, 2022.

\bibitem{zheltonozhskii2022contrast}
Evgenii Zheltonozhskii, Chaim Baskin, Avi Mendelson, Alex~M Bronstein, and
  Or~Litany.
\newblock Contrast to divide: Self-supervised pre-training for learning with
  noisy labels.
\newblock In {\em Proceedings of the IEEE/CVF Winter Conference on Applications
  of Computer Vision}, pages 1657--1667, 2022.

\bibitem{iscen2022learning}
Ahmet Iscen, Jack Valmadre, Anurag Arnab, and Cordelia Schmid.
\newblock Learning with neighbor consistency for noisy labels.
\newblock In {\em Proceedings of the IEEE/CVF Conference on Computer Vision and
  Pattern Recognition}, pages 4672--4681, 2022.

\bibitem{karim2022cnll}
Nazmul Karim, Umar Khalid, Ashkan Esmaeili, and Nazanin Rahnavard.
\newblock Cnll: A semi-supervised approach for continual noisy label learning.
\newblock In {\em Proceedings of the IEEE/CVF Conference on Computer Vision and
  Pattern Recognition}, pages 3878--3888, 2022.

\bibitem{smart2023bootstrapping}
Brandon Smart and Gustavo Carneiro.
\newblock Bootstrapping the relationship between images and their clean and
  noisy labels.
\newblock In {\em Proceedings of the IEEE/CVF Winter Conference on Applications
  of Computer Vision}, pages 5344--5354, 2023.

\bibitem{kingma2013auto}
Diederik~P Kingma and Max Welling.
\newblock Auto-encoding variational bayes.
\newblock {\em arXiv preprint arXiv:1312.6114}, 2013.

\bibitem{akrami2022robust}
Haleh Akrami, Anand~A Joshi, Jian Li, Serg{\"u}l Ayd{\"o}re, and Richard~M
  Leahy.
\newblock A robust variational autoencoder using beta divergence.
\newblock {\em Knowledge-Based Systems}, 238:107886, 2022.

\bibitem{Odaibo2019TutorialDT}
Stephen~G. Odaibo.
\newblock Tutorial: Deriving the standard variational autoencoder (vae) loss
  function.
\newblock {\em ArXiv}, abs/1907.08956, 2019.

\bibitem{mollah2010robust}
Md~Nurul~Haque Mollah, Nayeema Sultana, Mihoko Minami, and Shinto Eguchi.
\newblock Robust extraction of local structures by the minimum
  $\beta$-divergence method.
\newblock {\em Neural Networks}, 23(2):226--238, 2010.

\bibitem{mollah2006exploring}
Md~Nurul~Haque Mollah, Mihoko Minami, and Shinto Eguchi.
\newblock Exploring latent structure of mixture ica models by the minimum
  $\beta$-divergence method.
\newblock {\em Neural Computation}, 18(1):166--190, 2006.

\bibitem{kompass2007generalized}
Raul Kompass.
\newblock A generalized divergence measure for nonnegative matrix
  factorization.
\newblock {\em Neural computation}, 19(3):780--791, 2007.

\bibitem{ahmed2021relative}
Sara Atito~Ali Ahmed and Berrin Yanikoglu.
\newblock Relative attribute classification with deep-ranksvm.
\newblock In {\em Pattern Recognition. ICPR International Workshops and
  Challenges: Virtual Event, January 10--15, 2021, Proceedings, Part II}, pages
  659--671. Springer, 2021.

\bibitem{liu2015faceattributes}
Ziwei Liu, Ping Luo, Xiaogang Wang, and Xiaoou Tang.
\newblock Deep learning face attributes in the wild.
\newblock In {\em Proceedings of the IEEE international conference on computer
  vision}, pages 3730--3738, 2015.

\bibitem{sharif2014cnn}
Ali Sharif~Razavian, Hossein Azizpour, Josephine Sullivan, and Stefan Carlsson.
\newblock Cnn features off-the-shelf: an astounding baseline for recognition.
\newblock In {\em Proceedings of the IEEE conference on computer vision and
  pattern recognition workshops}, pages 806--813, 2014.

\bibitem{song2014exploiting}
Fengyi Song, Xiaoyang Tan, and Songcan Chen.
\newblock Exploiting relationship between attributes for improved face
  verification.
\newblock {\em Computer Vision and Image Understanding}, 122:143--154, 2014.

\bibitem{zhu2014multi}
Zhenyao Zhu, Ping Luo, Xiaogang Wang, and Xiaoou Tang.
\newblock Multi-view perceptron: a deep model for learning face identity and
  view representations.
\newblock {\em Advances in neural information processing systems}, 27, 2014.

\bibitem{rozsa2016facial}
Andras Rozsa, Manuel G{\"u}nther, Ethan~M Rudd, and Terrance~E Boult.
\newblock Are facial attributes adversarially robust?
\newblock In {\em 2016 23rd International Conference on Pattern Recognition
  (ICPR)}, pages 3121--3127. IEEE, 2016.

\bibitem{zhong2016face}
Yang Zhong, Josephine Sullivan, and Haibo Li.
\newblock Face attribute prediction using off-the-shelf cnn features.
\newblock In {\em 2016 International Conference on Biometrics (ICB)}, pages
  1--7. IEEE, 2016.

\bibitem{aly2018multi}
Sara~Atito Aly and Berrin Yanikoglu.
\newblock Multi-label networks for face attributes classification.
\newblock In {\em 2018 IEEE International Conference on Multimedia \& Expo
  Workshops (ICMEW)}, pages 1--6. IEEE, 2018.

\bibitem{8756609}
Vivek Sharma, Makarand Tapaswi, M.~Saquib Sarfraz, and Rainer Stiefelhagen.
\newblock Self-supervised learning of face representations for video face
  clustering.
\newblock In {\em 2019 14th IEEE International Conference on Automatic Face \&
  Gesture Recognition (FG 2019)}, pages 1--8, 2019.

\bibitem{2210.03853}
Yuxuan Shu, Xiao Gu, Guang-Zhong Yang, and Benny Lo.
\newblock Revisiting self-supervised contrastive learning for facial expression
  recognition, 2022.

\bibitem{1808.06882}
Olivia Wiles, A.~Sophia Koepke, and Andrew Zisserman.
\newblock Self-supervised learning of a facial attribute embedding from video,
  2018.

\bibitem{kinakh2021scatsimclr}
Vitaliy Kinakh, Olga Taran, and Svyatoslav Voloshynovskiy.
\newblock Scatsimclr: self-supervised contrastive learning with pretext task
  regularization for small-scale datasets.
\newblock In {\em Proceedings of the IEEE/CVF International Conference on
  Computer Vision}, pages 1098--1106, 2021.

\bibitem{alphabetagamma}
Andrzej Cichocki and Shun-ichi Amari.
\newblock Families of alpha- beta- and gamma- divergences: Flexible and robust
  measures of similarities.
\newblock {\em Entropy}, 12(6):1532--1568, 2010.

\bibitem{basu1998robust}
Ayanendranath Basu, Ian~R Harris, Nils~L Hjort, and MC~Jones.
\newblock Robust and efficient estimation by minimising a density power
  divergence.
\newblock {\em Biometrika}, 85(3):549--559, 1998.

\end{thebibliography}

\begin{appendices}
\section{Different Divergences}
\label{app:div}
The Kullback-Leibler (KL) divergence, a measure of the difference between two probability distributions, can be generalized by using a family of functions known as generalized logarithm functions or $\alpha$-logarithm,
\begin{equation}
    \log_{\alpha}(x) = \frac{1}{1-\alpha}(x^{1-\alpha} - 1)
\end{equation}
(for $x > 0$) which is a power function of x with power $1-\alpha$. The natural logarithm function is included in this family as a special case, where $\alpha \rightarrow 1$ \cite{alphabetagamma}.

By utilizing the concept of generalized logarithm functions, a family of divergences can be derived, which are known as the Alpha, Beta, and Gamma divergences. These divergences are extensions of the Kullback-Leibler (KL) divergence and can be used to measure the dissimilarity between two probability distributions in a more flexible way. Each of these divergences is defined using a different function and parameterization, allowing for different trade-offs between sensitivity and robustness. Especially Beta- and Gama- divergences are robust in respect to outliers for some values of tuning parameters, but Gama- divergence is a "super" robust estimation of some parameters in presence of outlier. Thus, beta divergence is more common choice for practical algorithms in literature, for example, robust PCA and clustering \cite{mollah2010robust}, robust ICA \cite{mollah2006exploring}, and robust NMF \cite{kompass2007generalized} and robust VAE \cite{akrami2022robust}.

\section{Beta-divergence Formulation}
\label{app:beta}
We consider a parametric model $p_\theta(X)$ with parameter $\theta$ and minimize the KL divergence between the two distributions (i.e. empirical distribution and probability distribution)\cite{akrami2022robust}:
\begin{equation}
    D_{KL}(p(X)||p_{\theta}(X)) = \int p(X) \log \frac{p(X)}{p_{\theta}(X)} dx
    \label{eq:kl}
\end{equation}
where  $p(X)=\frac{1}{N}\sum_{i=0}^N\delta(X,x^i)$ is the empirical distribution and its approximation to be converged. 
This is equivalent to minimizing \textit{maximum likelihood estimation}:
\begin{equation}
    arg\min_{\theta} \frac{1}{N}\sum_{i=1}^N \ln p_{\theta}(X) \nonumber
\end{equation}

Unfortunately, this formulation is sensitive to outliers because all data points contributes to the error with equal ratio. The density power divergence, or beta-
divergence, is a robust alternative to above formulation, proposed in \cite{basu1998robust}:

\begin{align}
    D_{\beta} (g||f) = - \frac{\beta+1}{\beta} \int g(x)(f(x)^{\beta} - 1) dx \nonumber + \int f(x)^{1+\beta} dx \nonumber
\end{align}

To motivate the use of the beta-divergence in \ref{eq:kl}, please note that minimizing the beta-divergence with empirical distribution yields:
\begin{eqnarray}
    0 = \frac{1}{N} \sum_{i=1}^N p_{\theta}(X)^\beta \frac{\partial}{\partial\theta}\ln p_{\theta}(X) \nonumber - \mathbf{E}_{p_{\theta}(X)} p_{\theta}(X)^\beta \frac{\partial}{\partial\theta} \ln p_{\theta}(X) \nonumber
\end{eqnarray}

\noindent where the second term assures the unbiasedness of the estimator and the first term is likelihood weighted according to the power of the probability density for each data point. Thus, the weights of the outliers will be much less then the major inliers, and the system will be more robust to noise.

Instead of using the empirical distribution, we can reduce the difference between two distributions from the same network:

\begin{eqnarray}
    &&D_{\beta}(p_{\theta}(Z_j|X_j)||p_{\theta}(Z_i|X_i)) =\nonumber\\ &&-\frac{\beta+1}{\beta} \int p_{\theta}(Z_j|X_j)(p_{\theta}(Z_i|X_i)^{\beta} - 1)dX + \int p_\theta(Z_i|X_i)^{\beta+1}dX \nonumber
\end{eqnarray}
where $p_{theta}(Z|X)$ are posterior distributions. This formulation is essential to the formulation of a robust self-supervised objective function in Eqn. \ref{eq:beta}.

\section{Derivation of Variational Objective Functions}
\label{app:variational}
The KL divergence between the two Gaussian distributions can be reduced to(i.e. empirical distribution and probability distribution) \cite{vcl}:
\begin{equation}
     -D_{KL} (q_{\theta}(z|x_i)||p(z)) = \log\left(\frac{\sigma_q}{\sigma_p}\right) - \frac{\sigma_q^2 + (\mu_q-\mu_p)^2}{2\sigma_p^2} + \frac{1}{2}
     \label{eq:mideq}
\end{equation}
\noindent From this formulation we will derive two variational objective terms in the paper.
\begin{itemize}
    \item Distribution Normalizing Loss \ref{app:norm}
    \item Distribution Similarity Loss by using Jensen-Shannon Divergence \ref{app:jsd}
\end{itemize}

\subsection{Normalizing Objective}
\label{app:norm}
\noindent We take the $\sigma_p=1$ and $\mu_p=0$ for Eq. \ref{eq:mideq}:
\begin{eqnarray}
    D_{KL} (q_{\theta}(z|x_i)||p(z)) &=& -\log(\sigma_q) + \frac{\sigma_q^2 + \mu_q^2}{2} - \frac{1}{2} \nonumber\\
    &=& -\frac{1}{2}\log(\sigma_q^2) + \frac{\sigma_q^2 + \mu_q^2}{2} - \frac{1}{2} \nonumber\\
    &=& - \frac{1}{2}\bigl[1 + \log(\sigma_q^2) - \sigma_q^2 - \mu_q^2\bigr]
\end{eqnarray}

\subsection{Distribution Similarity Objective using Jensen-Shannon Divergence}
\label{app:jsd}
\begin{align}
    \mathcal{L}_{a,k} &= JSD(q_1 || q_2) = \frac{1}{2} \big( D_{KL}(q_1 || m) + D_{KL}(q_2 || m)\big) \nonumber\\
    &\text{where} \hspace{3mm} m = \frac{1}{2}\left(q_1 + q_2\right)\nonumber
\end{align}
\noindent where the the sum of KL-divergences can be reduced to variational network output:
\begin{align}
    D_{KL}(q_1 || m) + D_{KL}(q_2 || m) = -\log\left(\frac{\sigma_{q_1}}{\sigma_{m}}\right) + \frac{\sigma_{q_1}^2 + (\mu_{q_1}-\mu_{m})^2}{2\sigma_{m}^2} - \frac{1}{2} &\nonumber\\ 
    -\log\left(\frac{\sigma_{q_2}}{\sigma_{m}}\right) + \frac{\sigma_{q_2}^2 + (\mu_{q_2}-\mu_{m})^2}{2\sigma_{m}^2} - \frac{1}{2} &\nonumber\\
    = -\left(\log(\sigma_{q_1}) -\log(\sigma_{m})\right) -\left(\log(\sigma_{q_2}) -\log(\sigma_{m})\right) &\nonumber\\
    + \frac{\sigma_{q_1}^2 + \sigma_{q_2}^2}{2\sigma_{m}^2} + \frac{(\mu_{q_1}-\mu_{m})^2}{2\sigma_{m}^2} + \frac{(\mu_{q_2}-\mu_{m})^2}{2\sigma_{m}^2} - 1 &\nonumber\\
    = -\left(\log(\sigma_{q_1}) -\log(\sigma_{m})\right) -\left(\log(\sigma_{q_2}) -\log(\sigma_{m})\right) &\nonumber\\
    + \frac{(\mu_{q_1}-\mu_{m})^2 + (\mu_{q_2}-\mu_{m})^2}{2\sigma_{m}^2} &\nonumber
\end{align}

\section{Ablation Studies for Objective Functions and Hyper-parameters}
\label{app:ablation}
\begin{table}[thb]

\centering
\caption{Ablation Studies: First table showing the accuracy obtained with and without the objective components. Second table is the hyper-parameter response of the system.}
\begin{tabular}{l|c|c|c|c|c}\midrule
Objective Comp. & beta-NT-Xent      & Dist. Sim.       & Dist. Norm.      &  CelebA  & YFCC-CelebA \\\hline
Accuracy             &         yes       &        no        &        no        &  86.19\%   & 85.93\% \\
                     &         yes       &        yes       &        no        &  88.42\%   & 87.34\% \\
                     &         yes       &        no        &        yes       &  88.54\%   & 87.87\% \\\hline                
\end{tabular}
\end{table}
\begin{table}[thb]

\centering
\begin{tabular}{l|cc|c|cc}
&CelebA &YFCC-CelebA & &CelebA &YFCC-CelebA \\\hline
temp. (T) &Acc. &Acc. \%1 &beta (T=0.07) &Acc. &Acc. \%1 \\\hline
0.07 &\textbf{89.20} &\textbf{87.49} &0.0010 &88.96 & 87.65 \\
0.1 &88.87 &87.26 &0.0050 &\textbf{89.23} & \textbf{88.12} \\
0.2 &88.69 &86.79 &0.0100 &89.17 & 88.01 \\
\bottomrule
\end{tabular}
\end{table}
\end{appendices}

\end{document}